\documentclass[runningheads]{llncs}
\usepackage{graphicx}
\usepackage{caption}
\usepackage{hyperref}
\begin{document}
\title{Automatic Identification of Traditional Colombian Music Genres based on Audio Content Analysis and Machine Learning Techniques}
\titlerunning{Automatic identification of traditional Colombian music}
%
\author{
Diego A. Cruz, 
Sergio S. Lopez, 
Jorge E. Camargo 
}


%
%

\institute{Universidad Nacional de Colombia, Bogota, Colombia\\
\url{http://www.unsecurelab.org} \\
\email{\{diacruzmo,sslopezm,jecamargom\}@unal.edu.co}
}

%
\maketitle              
\begin{abstract}
Colombia has a diversity of genres in traditional music, which allows to express the richness of the Colombian culture according to the region. This musical diversity is the result of a mixture of African, native Indigenous, and European influences. Organizing large collections of songs is a time consuming task that requires that a human listens to fragments of audio to identify genre, singer, year, instruments and other relevant characteristics that allow to index the song dataset. This paper presents a method to automatically identify the genre of a Colombian song by means of its audio content. The method extracts audio features that are used to train a machine learning model that learns to classify the genre. The method was evaluated in a dataset of 180 musical pieces belonging to six folkloric Colombian music genres: Bambuco, Carranga, Cumbia, Joropo, Pasillo, and Vallenato. Results show that it is possible to automatically identify the music genre in spite of the complexity of Colombian rhythms reaching an average accuracy of 69\%.

\end{abstract}
\section{Introduction}

Traditional Colombian music has a clear expression of the country's culture, achieving through its diffusion to share some characteristic of our society. Colombian musical diversity is the result of a mixture of African, native Indigenous, and European influences. The popularity of the Colombian music genres depends on the region. For instance, in the Andean region the most popular genres are Bambuco, Pasillo and Carranga; in the Orinoquia region genres such as the Joropo, Contrapunteo and Pajarillo are in the most popular; in the Caribbean region Cumbia, Vallenato and Porro are the most representative genres; in the Insular region genres such as Reggae, Pasillo Isle\~{n}o, Vals Isle\~{n}o are in the main genres in the Colombian islands; in the Pacific region Currulao, Patacorée and Mekerule are in the most popular.

One of the main influences in the central region of Colombia (the Andean Region) was the Waltz genre, which typically sounds one chord per measure, and the accompaniment style particularly associated is to play the root of the chord on the first beat, the upper notes on the second and third beats.

Companies nowadays use music classification, by means of recommendation systems like "Spotify" or simply as a product like "Shazam". The traditional musical genres of Colombia are not widely recognized globally, and in some cases, they have lost popularity and remain exclusively in the regions to which they belong culturally. Identify the musical genres is the first step to exalt and make known more easily both internally and globally this traditional music. Indexing in music information retrieval systems is a very important task to allow search in a music dataset. Machine learning techniques have proved to be successful in the analysis of trends and patterns of music, which although over time and the study of them have increased research on their application in music, there are no enough research focused on the identification of musical genres.

This paper proposes an automatic method to identify the genre of Colombian music. Particularly, we focused on some popular Colombian genres: Bambuco, Carranga, Cumbia, Joropo, Pasillo, and Vallenato. Up to the best of our knowledge this is the first attempt to automatically classify these Colombian genres. The proposed model can be used in the construction of new music information retrieval systems and music recommendation systems to allow the access to Colombian music, which in many cases is inaccessible with current technology. We want to contribute also in the continuity of cultural heritage in these days when traditional Colombian music is increasingly forgotten in the new generations.

The rest of the paper is organized as follows: Section 2 presents related work; Section 3 presents the proposed model; in Section 4 results are presented; and Section 5 concludes the paper.

\section{Related Work}

The classification of musical genres is a field that has always been of great interest in the scientific community in the last times with the application of supervised machine learning techniques, such as Gaussian Mixture model \cite{Gaussian2013} and k-nearest neighbour classifiers \cite{KNearest2014}. There have been several works that seek to refine more and more the methods to obtain better classifiers. In Hareesh Bahuleyan \cite{machinelearning2018},  authors use conventional machine learning models like Logistic Regression, Random Forests and Gradient Boosting which are trained to classify the audio pieces. Tao Feng \cite{deep2014} studies the pre-trained algorithms such as auto-encoders and restricted Boltzmann machine. 
Thiruvengatanadhan \cite{classification2018} applies a technique that uses support vector machines (SVM) to classify songs based on features using Mel Frequency Cepstral Coefficients (MFCC).

\section{Proposed Method}
This section presents the proposed method, which is composed of a song feature extraction process and the training of machine learning models that learn from the audio content.

\subsection{Feature Extraction}

Music information retrieval is a science that is responsible for the retrieval of the musical information, currently we can find applications like: the classification of musical genres, music transcription, speech recognition, among others. For this work we have used "Librosa"\cite{librosa2019}, a Python module specialized in performing this task. The features chosen in this investigation are the following ones:

\paragraph{Spectogram:\newline}
It is a visual representation of the frequency spectrum in a signal, which varies with time (see Figure \ref{fig3}). A common format is an image that indicates: on the vertical axis the frequency, on the horizontal axis time and a third dimension with the amplitude of a particular frequency at a given moment, represented by the intensity of the color.

\begin{figure}
\centering
\includegraphics[scale=0.3]{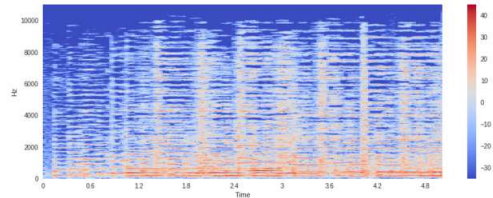}\\
\caption{Spectrogram of a Pasillo song} \label{fig1}
\end{figure}

\paragraph{Spectral Centroid:\newline}
This property tells us where the mass center is located in a spectrogram and is obtained with the weighted average of the frequencies (see Figure \ref{fig3}). It helps to predict the "brightness" in a sound, so it is very useful when measuring the "timbre" in an audio.

\begin{figure}
\centering
\includegraphics[scale=0.3]{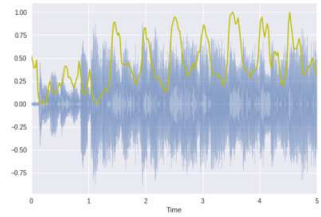}
\caption{Spectral Centroid} \label{fig2}
\end{figure}

\paragraph{Chroma Features:\newline}
It is strongly related to the 12 semitones in the music, it is a powerful tool for audio analysis that has tones that can be categorized in a significant way, one of its properties allows to capture the harmonic and melodic characteristics (see Fig. 3).

\begin{figure}
\centering
\includegraphics[scale=0.3]{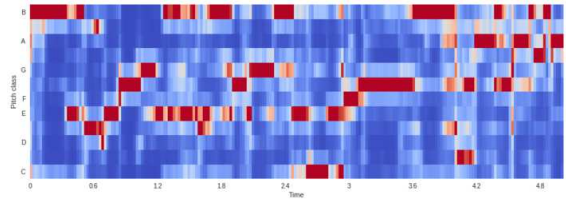}
\caption{Chroma Features} \label{fig3}
\end{figure}

\paragraph{Zero Crossing Rate:\newline}
It is the amount of sign changes that a signal experiences, in other words, it is the number of times the signal changes its value, from positive to negative and vice versa. It is used to measure the amount of noise in a signal.

\paragraph{Mel-Frequency Cepstral Coefficients:\newline}
It is small set of characteristics that concisely describe the general way of a spectral envelope. It is widely used in the retrieval of musical information, to obtain audio similarity measurements and in the classification of genres.

\paragraph{Spectral Rolloff:\newline}
It is a measure of the shape of a signal, it indicates the frequency in Hz that is below a percentage of the total spectral energy.

\begin{figure}
\centering
\includegraphics[width=0.6\textwidth]{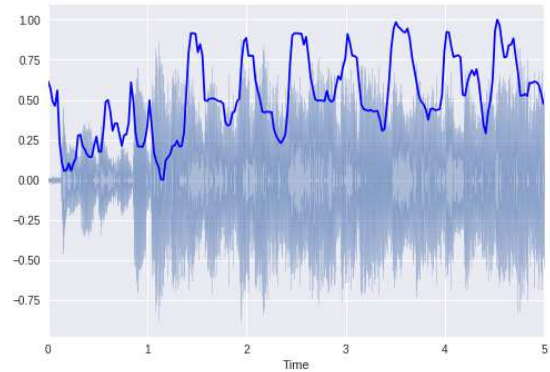}
\caption{Spectral Rolloff} \label{fig9}
\end{figure}

\subsection{Model Learning}
The experimental section is realized through Co-laboratory, a free Jupyter Notebook environment that does not require configuration and that is completely executed in the cloud. Both supervised and unsupervised classification will be used.

We will use two methods of supervised classification, which are Random Forest\cite{randomforest2018} with the help of the Sci-kit Learn library (Sklearn) and the training of a neural network\cite{retrieval2016} of 4 layers with the help of the Keras library. Validation partitions will be created to verify that the training of the models does not fall into underfitting or overfitting. Therefore, it will also be analyzed if it is necessary for our data set to make a reduction in the number of characteristics or not. The performance of the method will be evaluated using metrics such as accuracy, error and accuracy, recall and score per class.

For the unsupervised classification, Clustering\cite{clustering2015} will be used with the help of centroid-based algorithms (k-means) to see how the groups will be distributed in 3 configurations; first with the original data, then with a reduction of dimensionality using PCA and finally applying t-SNE.

\paragraph{Supervised Classification with Random Forest\newline}
The first method that we use is Random Forest, it is a very accurate learning algorithm that consists of a combination of prediction trees, each tree depends on the values of a randomly tested vector independently and with the same distribution for each of these, building a long collection of uncorrelated trees and then averaging them.
To apply this classifier, before training it, an analysis of its complexity is performed for different values of estimators, this with the objective of finding the model with the best relation between training error and generalization error (see Fig. 4). Choosing the value of \(2^6\).

\paragraph{Supervised Classification with Neural Network\newline}
The multilayer perceptron (MLP) is a special type of neural network\cite{neuralnetwork2016} in which several layers of perceptrons are stacked. It is also called Feedforward neural network. The multilayer perceptron is motivated by the little ability of the simple perceptron to model nonlinear functions.
For our neural network, we used the Keras library, which allows us to define the base model and add layers as required, in our case our network is built with 4 layers of 256, 128, 64 and 6 neurons respectively.

\begin{figure}
\centering
\includegraphics[width=0.6\textwidth]{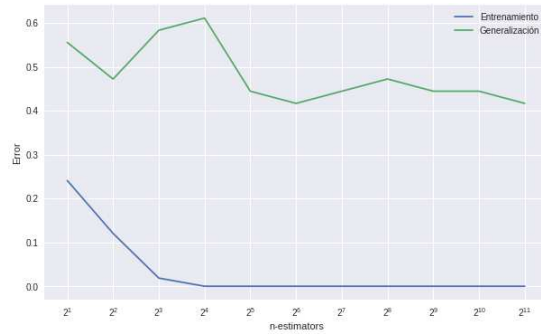}
\caption{Random Forest} \label{fig4}
\end{figure}

\paragraph{Clustering Analysis\newline}
The objective of Clustering is to group physical or abstract objects in classes of similar objects, it is an unsupervised task, because we do not know how to classify our objects, so the algorithm will only pass the data of the set and not its labels.

\paragraph{Dimensionality Reduction with PCA\newline}
Principal component analysis is a technique used to describe a set of data in terms of new uncorrelated variables called "components". The components are ordered by the amount of original variance, so the technique is useful to reduce the dimensionality of a set of data. This technique is used mainly in exploratory data analysis and in the construction of predictive models. For our particular case when evaluating our data set, which contained 26 different features, we were able to obtain the following accumulated variance graph (see Fig. 5).

\begin{figure}
\centering
\includegraphics[width=0.6\textwidth]{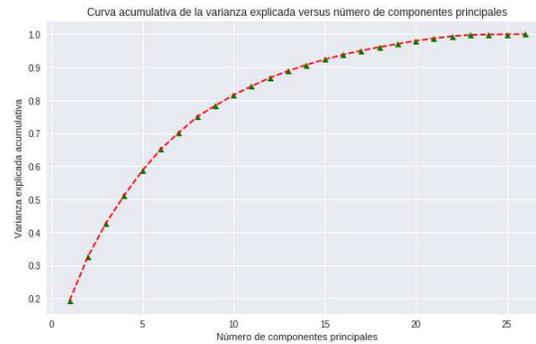}
\caption{Accumulated Variance} \label{fig5}
\end{figure}

Since our objective is to reduce these characteristics without losing valuable information, we decided to use the principal components analysis to transform our data set into one with 10 components, since its variance is around 82\% and more than half of the characteristics are being eliminated.

\paragraph{Dimensionality Reduction with t-SNE\newline}
It is an automatic learning algorithm for visualization developed by Laurens Van Der Maaten and Geoffrey Hinton. It is a non-linear dimensionality reduction technique adapted to embed high-dimensional data for visualization in a low-dimensional space in two or three dimensions. Specifically, it models each high-dimensional object by a point of two or three dimensions in such a way that similar objects are modeled by nearby points and different objects are modeled by distant points with high probability. The algorithm naturally uses the Euclidean distance, but this can be modified in the metrics of the same, in our case we use Minkowski.

\section{Results}
This section presents the obtained results of the conducted classification experiments.
\subsection{Classification Results}
Table \ref{tab1} presents the confusion matrix obtained using the Random Forest classifier. Note that the Cumbia genre is the most difficult for this classifier.

        \begin{table}
        \caption{Confusion Matrix for the random forest classifier}\label{tab1}
        \centering
        \includegraphics[width=0.75\textwidth]{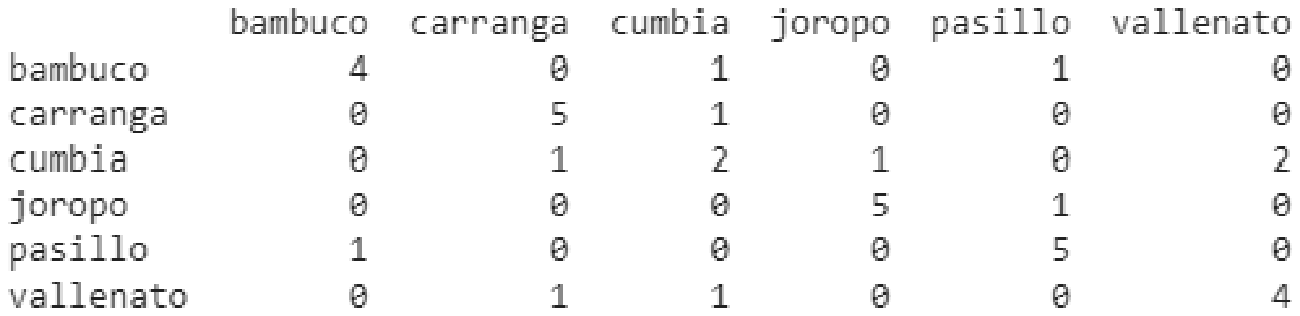}
        \end{table}	
        
In Table \ref{tab2} the confusion matrix for the neural network classifier is presented. It is worth noting that Bambuco genre is perfectly classified.

	    \begin{table}
        \caption{Confusion Matrix for the ANN classifier}\label{tab2}
        \centering
        \includegraphics[width=0.75\textwidth]{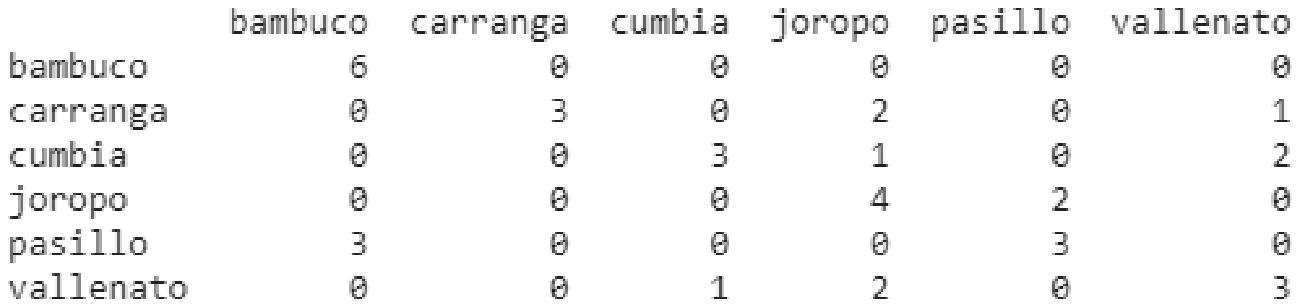}
        \end{table}	
        
In Table \ref{tab3} a comparison of random forest and ANN is performed. The performance metrics analyzed are accuracy, error, precision, recall and f1-score. The ANN obtained the highest classification performance in terms of accuracy, reaching 69\%.

	    \begin{table}
        \caption{Random Forest vs Neural Network.}\label{tab3}
        \centering
        \includegraphics[width=0.75\textwidth]{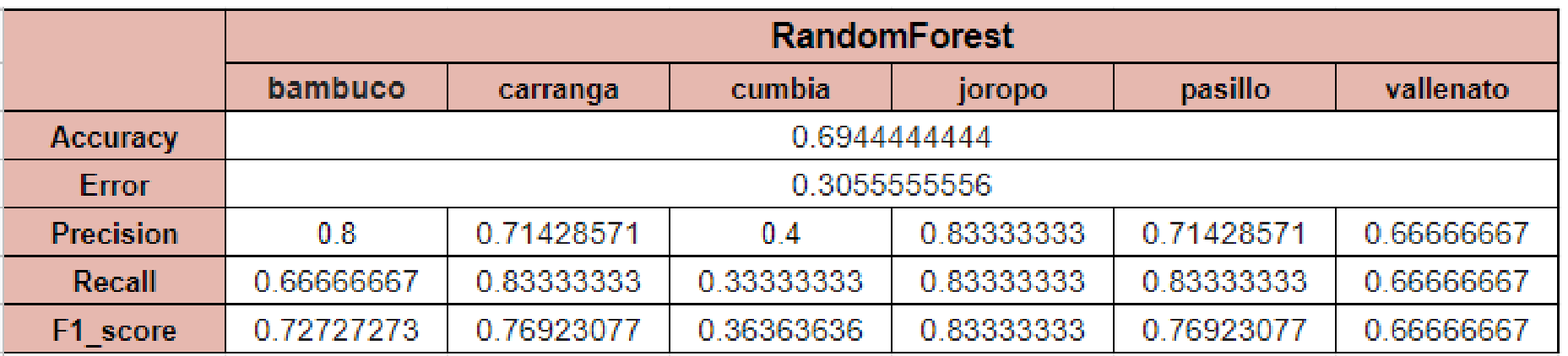}
        \includegraphics[width=0.75\textwidth]{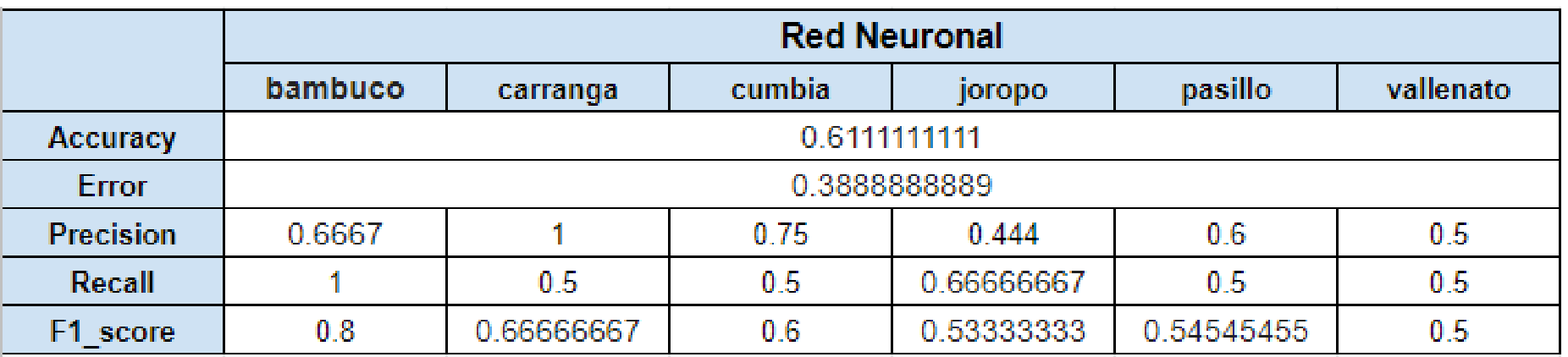}
        \end{table}

\subsection{2D Visualization}

Figure \ref{fig7} presents a 2D visualization of the songs using all the extracted features. The highlighted circles represent the centroids of the clusters found with k-means. Figure \ref{fig8} shows a visualization using PCA and Figure \ref{fig8} using t-SNE. In this we found that the best number of clusters is 6 using the coefficient silhouette (sc) analysis. In each Figure the inertia score is reported.

it is important to note that reducing the dimensionality of the vector that represents a song produces a more compact feature representation. In this case, results show that t-SNE generates the best visualization of the complete song dataset.

        \begin{figure}
        \centering
        \includegraphics[width=0.6\textwidth]{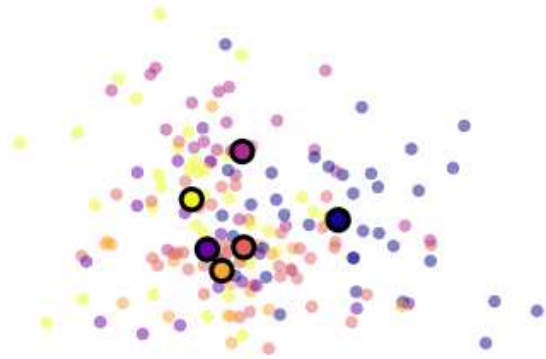}
        \caption{2D visualization of the song dataset using all the extracted features. Clusters = 6, inertia = 3227, sc = 0.096} \label{fig6}
        \end{figure}

        \begin{figure}
        \centering
        \includegraphics[width=0.6\textwidth]{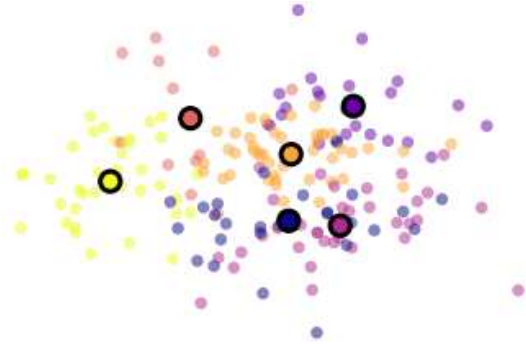}
        \caption{2D visualization of the song dataset using PCA to reduce the dimensionality. Clusters = 6, inertia = 2384, sc = 0.127} \label{fig7}
        \end{figure}	
        
        \begin{figure}
        \centering
        \includegraphics[width=0.6\textwidth]{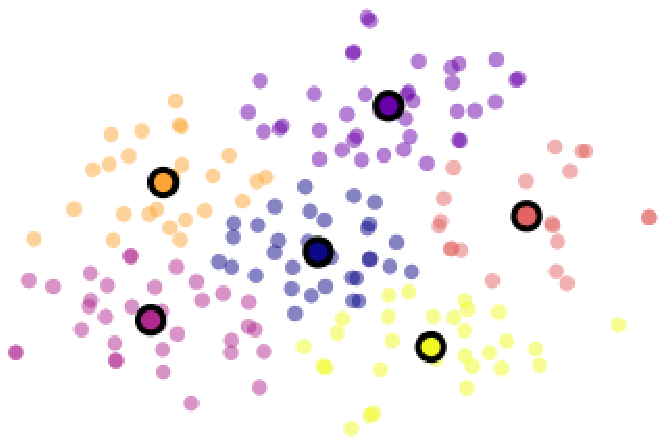}
        \caption{2D visualization of the song dataset using t-SNE to reduce the dimensioanlity. Clusters = 6, inertia = 2477, sc = 0.387} \label{fig8}
        \end{figure}	

On the unsupervised classification side, it is very clear to observe the improved in the visualization of the different groups or clusters formed after the application of dimensionality reduction methods such as PCA and t-SNE; being the latter one that better allows us to observe the separation between the clusters. Possibly by increasing the number of songs in the data set, the use of t-SNE will be the next step if an unsupervised classification by means of clusters is desired.

After analyzing the results of this work, the next step to follow is the feeding of our data set until we obtain a minimum of 1000 songs in order to have more diversity of data, which will allow us train better our model , in addition to this, the analysis of other features that could be important for the classification and add them to the proposed model.

In the other hand, if we see the Colombian`'s traditional genres we could find that many of them have subgenres that also represent cultures of different regions, for this we think that this work can be applied to a specific genre to classify this sub-genres, for example in the Vallenato we could find the Son, Paseo, Merengue and Puya; in Joropo we could find Contrapunteo, Pasaje, Tonada, Golpe llanero and Copla; and in Carranga the subgenres Rumba and Merengue. Using this we can improve our model to specify not only the genre but also the sub-genre of the song.

We also would like to obtain information about the instruments that are identified in each song, because also involve the culture of the regions and could be an important feature that helps in classification.
  
\section{Conclusion and Future Work}
This paper presented a method based on two classification techniques (supervised and unsupervised). The first technique was the Random Forest, which had the highest performance with 8\% more success than the. This may be due to the previous evaluation of the relationship between training error and generalization error; with which we reduce the probability that when training this classifier is so complex as to remember the particularities of the training set (about adjustment/overfitting) or so flexible so as not to model the variability of the data (subfitting/underfitting); an aspect that we only manage with the validation partition in the training of the neural network.

Also we can highlight with the help of metrics and the confusion matrix that the Cumbia is the most difficult of identify musical genre for Random Forest and too one of the least accurate in the neural network, so we can conclude that it is the Colombian genre that is more difficult to classify using these methods.

%
%
%
%
\bibliographystyle{ieeetr}
\bibliography{main}

\end{document}